\documentclass[applsci,article,accept,pdftex,moreauthors]{Definitions/mdpi} 

\usepackage{booktabs, multirow, placeins, float, graphicx, indentfirst, mathrsfs, amsmath, tabularx, adjustbox, array, pifont, subcaption, caption, amssymb, algorithm, algpseudocode, url, amsfonts, cancel, titlesec, lipsum, xcolor}

\firstpage{1} 
\pubvolume{1}
\issuenum{1}
\articlenumber{0}
\pubyear{2024}
\copyrightyear{2024}
\externaleditor{Academic Editor: Andrea Prati}
\datereceived{18 January 2024} 
\daterevised{16 February 2024} 
\dateaccepted{17 February 2024} 
\datepublished{ } 
\hreflink{https://doi.org/} 

\Title{VTG-GPT: Tuning-Free Zero-Shot Video Temporal Grounding with {GPT}} 

\TitleCitation{VTG-GPT: Tuning-Free Zero-Shot Video Temporal Grounding with GPT}

\Author{
{Yifang Xu} 
 $^{1}$\orcidA{}, Yunzhuo Sun $^{2}$, Zien Xie $^1$, Benxiang Zhai $^1$ and Sidan Du $^{1,}$*
}

\AuthorNames{Yifang Xu, Yunzhuo Sun, Zien Xie, Benxiang Zhai, and Sidan Du}

\AuthorCitation{{Xu, Y.;} 
 Sun, Y.; Xie, Z.; Zhai, B.; Du, S.}

\address{%
$^{1}$ \quad School of Electronic Science and Engineering, Nanjing University,
 {Nanjing 210093,} 
  China; {xyf@smail.nju.edu.cn (Y.X.); xze@smail.nju.edu.cn (Z.X.); zbx@smail.nju.edu.cn (B.Z.)} 
\\
$^{2}$ \quad School of Physics and Electronics, Hubei Normal University, {Huangshi 435002}, China; {sunyunzhuo98@gmail.com}\\
}

\corres{Correspondence: coff128@nju.edu.cn}

\abstract{Video temporal grounding (VTG) aims to locate specific temporal segments from an untrimmed video based on a linguistic query. Most existing VTG models are trained on extensive annotated video-text pairs, a process that not only introduces human biases from the queries but also incurs significant computational costs. To tackle these challenges, we propose VTG-GPT, a GPT-based method for zero-shot VTG without training or fine-tuning. To reduce prejudice in the original query, we employ Baichuan2 to generate debiased queries. To lessen redundant information in videos, we apply MiniGPT-v2 to transform visual content into more precise captions. Finally, we devise the proposal generator and post-processing to produce accurate segments from debiased queries and image captions. Extensive experiments demonstrate that VTG-GPT significantly outperforms SOTA methods in zero-shot settings and surpasses unsupervised approaches. More notably, it achieves competitive performance comparable to supervised methods. The code is available on \url{https://github.com/YoucanBaby/VTG-GPT}. 
}


\begin{document}

\section{Introduction}
\label{sec:intro}

\begin{figure}[H]
 
\begin{adjustwidth}{-\extralength}{0cm}
\centering 
 \includegraphics[width=.95\linewidth]{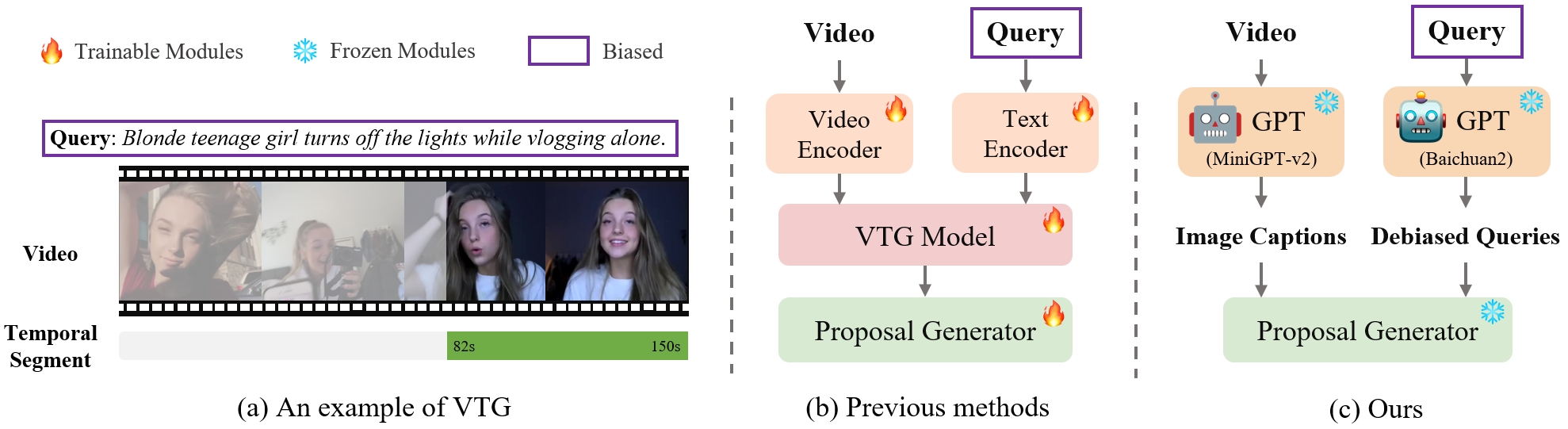}
\end{adjustwidth}
  \caption{(\textbf{a}) {An} 
 illustrative example of a video temporal grounding (VTG) task. (\textbf{b}) Previous methods require training for all modules. (\textbf{c}) Our proposed VTG-GPT operates without any training or fine-tuning. Moreover, it employs GPT to reduce bias in human-annotated {queries.} 
  }
  \label{Fig1}
\end{figure}

Given a linguistic query, video temporal grounding (VTG) aims to locate the most relevant temporal segments from an untrimmed video, each containing a start and end timestamp. An illustrative example of VTG is shown in Figure \ref{Fig1}a. This task \cite{MomentDETR-2021, Zero-shot-VMR-2023} has numerous practical applications in daily life, such as how it can help video platform users easily skip to relevant portions of a video. The field of natural language has witnessed a significant leap forward with the advent of GPT-4 \cite{ChatGPT4-2023}. This development has spurred the rise of large language models (LLMs) such as LLaMA \cite{LLaMa-2-2023} and Baichuan2 \cite{Baichuan-2-2023}. Concurrently, GPT-based (Generative Pre-trained Transformer) models like MiniGPT4 \cite{MiniGPT-v2} and LLaVA \cite{LLaVA-2023} have made significant strides in vision and multimodal applications. A recent work, LLaViLo~\cite{LLaViLo-2023}, reveals that training adapters alone can effectively leverage the video understanding capabilities of LLMs. However, this method requires designing a sophisticated fine-tuning strategy specifically for VTG, thereby introducing additional computing costs.

\begin{figure}[H]

\begin{adjustwidth}{-\extralength}{0cm}
\centering 
  \includegraphics[width=1.1\textwidth]{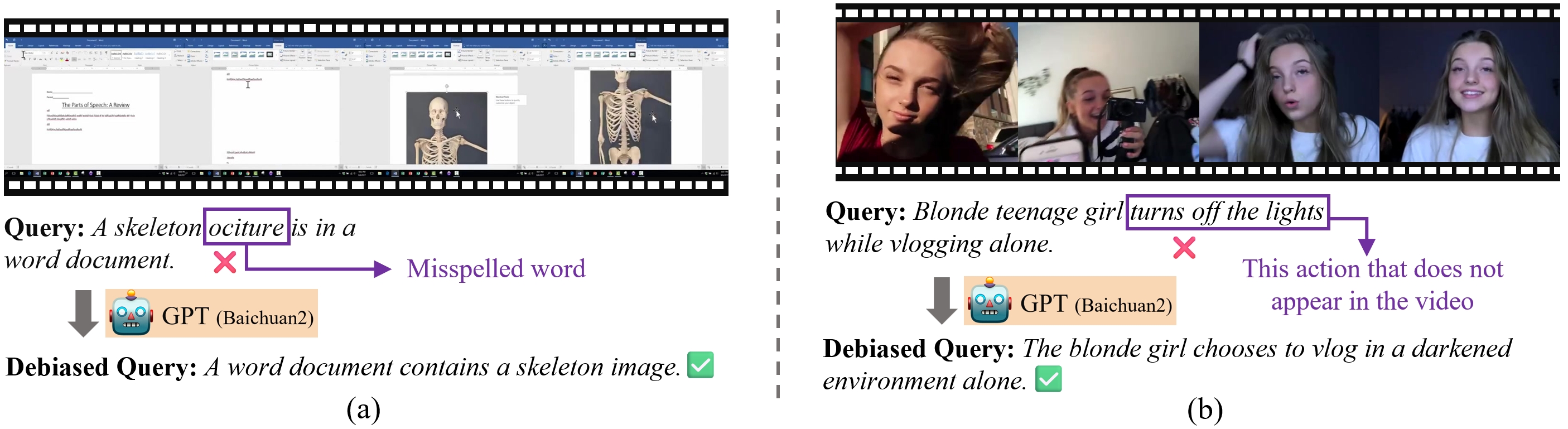}
\end{adjustwidth}
  \caption{{Human} 
 biases in ground-truth queries arise from (\textbf{a}) misspelled words and (\textbf{b}) incorrect descriptions. Our approach effectively mitigates these biases by leveraging GPT to optimize raw~queries.}
  \label{Fig2}
\end{figure}

Existing VTG methods \cite{MomentDETR-2021, UMT-2022, MH-DETR-2023, QD-Net-2023, VDI-2023} primarily adopt supervised learning, which demands massive training resources and numerous annotated video-query pairs, as illustrated in Figure \ref{Fig1}b. However, developing datasets for VTG is time-consuming and expensive; for instance, Moment-DETR \cite{MomentDETR-2021} spent 1455 person-hours and USD 16,600 to create the QVhighlights dataset. Furthermore, ground-truth (GT) queries often contain human biases, such as (1)~Bias from erroneous word spellings, as depicted in Figure \ref{Fig2}a. The misspelled word ``\textit{{ociture}
}'' in original query would be tokenized by language models into ``\textit{{o}}'', ``\textit{{cit}}'', ``\textit{{ure}}'', leading to model misunderstanding; (2) Bias due to incorrect descriptions, as shown in Figure \ref{Fig2}b. Here, the action ``\textit{{turns off the lights}}'' mentioned in the query does not occur in the video.

\vspace{-6pt}

In this paper, we propose a tuning-free zero-shot method named VTG-GPT to address the above issues. As shown in Figure \ref{Fig1}c, VTG-GPT completely satisfies zero-shot settings, adopting a direct feed-forward approach without training or fine-tuning. To minimize biases arising from human-annotated queries, we employ Baichuan2 \cite{Baichuan-2-2023} to rephrase the original query and obtain debiased queries. As illustrated in Figure \ref{Fig2}, the erroneous word ``\textit{{ociture}}'' in query (a) has been accurately revised to ``\textit{{image}}'', and the non-existent action ``\textit{{turn off the lights}}'' in query (b) has been effectively refined to ``\textit{{a darkened environment}}''. Furthermore, considering that videos inherently contain more redundant information than text, and inspired by the human approach to understanding video linguistically, we apply MiniGPT-v2 \cite{MiniGPT-v2} to transform visual content into more precise textual descriptions. Finally, to generate accurate temporal proposals, we design a proposal generator that models debiased queries and image captions in the textual domain. In summary, our main contributions include:

{(}
1) To the best of our knowledge, we are the first zero-shot method to utilize GPT on VTG without training or fine-tuning.

{(}2) We present a novel framework, VTG-GPT, which effectively leverages GPT to mitigate human prejudice in annotated queries. Furthermore, VTG-GPT distinctively models debiased queries and video content within the linguistic domain to generate temporal segments.

{(}3) Comprehensive experiments demonstrate that VTG-GPT significantly surpasses SOTA (State-of-the-Art) methods in zero-shot settings. More importantly, this method achieves competitive performance comparable to supervised methods.

\section{Related Work}
\label{sec:related_work}

\subsection{Video Temporal Grounding}
For fully-supervised VTG, prior works \cite{MomentDETR-2021, UMT-2022, MH-DETR-2023, QD-Net-2023, UniVTG-2023, jangKnowingWhereFocus2023, 2D-TAN-2020, GPTSee-2023} typically employ encoders to extract visual and textual features, followed by designing a VTG model (e.g., transformer encoder-decoder) to interact and align two modalities, as depicted in Figure \ref{Fig1}b. UniVTG~\cite{UniVTG-2023} designs a multi-modal and multi-task learning pipeline, undergoing pre-training or fine-tuning on dozens of datasets. To accelerate the training convergence of VTG, GPTSee~\cite{GPTSee-2023} introduces LLMs to generate prior positional information for the transformer decoder. However, these supervised approaches inevitably rely on extensive human-annotated data and training resources. To alleviate the dependence on annotations, PSVL~\cite{PSVL-2021}, DSCNet \cite{DSCNet-2022}, and Gao et al. \cite{gaoLearningVideoMoment2022} propose unsupervised frameworks that employ clustering to generate pseudo queries from video features. Similarly, PZVMR \cite{PZVMR-2022} and Kim et al. \cite{LFT-VG-2023} leverage CLIP \cite{CLIP-2021} for pseudo query generation. Yet, the above unsupervised methods unavoidably introduce biases from mismatched video-query pairs. {In this paper, we adhere to the definitions of unsupervised and zero-shot settings as discussed by Luo et~al. \cite{Luo2023ZeroShotVM}, classifying these approaches \cite{PSVL-2021, PZVMR-2022, LFT-VG-2023} as unsupervised.}

{To avoid any training or fine-tuning of the model,} 
Diwan et al. \cite{Zero-shot-VMR-2023} design the first zero-shot framework utilizing CLIP, but its reliance on shot transition detectors for obtaining temporal segments limits performance. Considering that CLIP (InternVideo \cite{InternVideo-2022}) pre-trained on 400 M image-text (12 M video-text) pairs can align visual and textual inputs in a shared feature space, Luo et al. \cite{Luo2023ZeroShotVM} develop a bottom-up pipeline to leverage the capabilities of vision-language models. 
{Wattasseril et al. \cite{wattasseril2023zero} employ the sparse frame-sampling strategy and BLIP2 \cite{li2023blip} to reduce the computational cost of inference.}
However, these zero-shot methods \cite{Zero-shot-VMR-2023, Luo2023ZeroShotVM, wattasseril2023zero} tend to generate redundant video features, introducing new biases that impair model performance. A recent study \cite{MAE-2022} found that masking over 75\% of the input images can effectively train large self-supervised models. Moreover, SeViLA \cite{SeViLA-2023} demonstrates that transforming visual signals into textual representations significantly reduces redundant information, thereby boosting performance in tasks such as video question answering and VTG.

\subsection{Generative Pre-Trained Transformer}
The groundbreaking success of GPT-4 \cite{ChatGPT4-2023} in the language domain has led to the development of a series of open-source LLMs \cite{WizardLM-2023, LLaMA-2023, LLaMa-2-2023, Baichuan-2-2023}. Baichuan2 \cite{Baichuan-2-2023}, containing 7 billion parameters and 2.6 trillion tokens, excels in vertical domains such as technology and daily conversation. MiniGPT4 \cite{MiniGPT4-2023, MiniGPT-v2} introduces a large multi-modal model (LMM) based on GPT, adept at performing visual-linguistic tasks like image captioning and visual question answering. Recent studies demonstrate that leveraging GPT models effectively reduces prejudice originating from ground truth labels, while simultaneously enhancing model performance in zero-shot multimodal tasks. This advancement is particularly notable in areas such as relation detection and information extraction, showcasing the robust generalization capabilities of GPT in these complex scenarios. To further capitalize on GPT's capabilities in video understanding, LLaViLo \cite{LLaViLo-2023} designs specialized adapters for VTG, but this method still necessitates model training. To overcome these limitations, this paper proposes a novel zero-shot VTG pipeline aiming to eliminate human biases from GT queries while fully harnessing the visual comprehension capabilities of GPT, achieving a tuning-free~framework.

\section{Our Method}
\label{sec:method}
In this section, we first formulate the VTG task and then present the overall architecture of our VTG-GPT. Subsequently, we provide details of each module in the model.

\subsection{Overview}
\label{sec:overview}
\begin{figure}[H]
    \hspace*{-3.5cm}
    \includegraphics[width=1.2\linewidth]{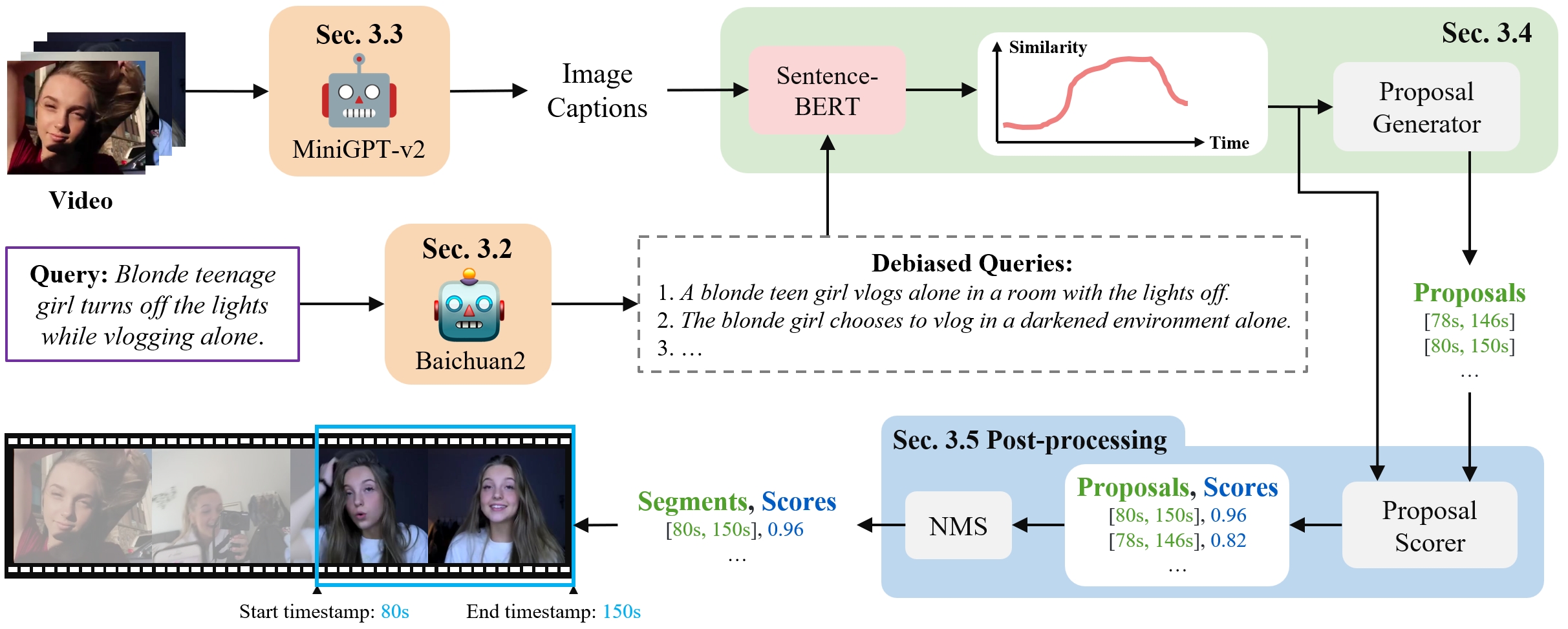}
    \caption{An overview of our proposed VTG-GPT, the framework contains four key phases: query debiasing (Section \ref{sec:query_debiasing}), image captioning (Section \ref{sec:image_captioning}), proposal generation (Section \ref{sec:proposal_generation}), and post-processing (Section \ref{sec:post_processing}).}
    \label{Fig3}
\end{figure}

Given an untrimmed video $V \in \mathbb{R}^{N_v \times H \times W \times 3}$ consisting of $N_v$ frames and a natural language query $T \in \mathbb{R}^{L_t}$ formed by $L_t$ words, the objective of video temporal grounding (VTG) is to precisely identify time segments [$t_s$, $t_e$] $\in \mathbb{R}^{N_s \times 2}$ in $V$ that semantically correspond to $T$, where each segment starts at timestamp $t_s$ and ends at timestamp $t_e$. The overview of our proposed VTG-GPT is illustrated in Figure \ref{Fig3}.

The core aim of VTG-GPT is to implement a tuning-free framework while reducing human bias in the process. To this end, our first step is employing Baichuan2 (\mbox{Section \ref{sec:query_debiasing}}) to refine raw query $T$, resulting in debiased queries $Q \in \mathbb{R}^{N_q \times L_t}$. Then, we leverage MiniGPT-v2 (Section \ref{sec:image_captioning}) to convert visual content in each frame into image captions $C \in \mathbb{R}^{N_v \times L_c}$, effectively reducing redundant information in video $V$. In Section \ref{sec:proposal_generation}, we compute similarity scores $S_s \in \mathbb{R}^{N_q \times N_v}$ between $Q$ and $C$, which is to say, in the linguistic domain via Sentence-BERT assess query-frame correlation. Following this, a proposal generator is designed to yield temporal proposals $P \in \mathbb{R}^{N_p \times 2}$. Finally, in the post-processing stage (Section \ref{sec:post_processing}), we calculate final scores $S_f \in \mathbb{R}^{N_p}$ for each proposal while removing excessively overlapping proposals to produce predicted segments $Seg \in \mathbb{R}^{N_s \times 2}$.

\subsection{Query Debiasing}
\label{sec:query_debiasing}
Mitigating biases in ground-truth queries represents a crucial and challenging problem for VTG, as these biases often originate from inherent human subjectivity. Such biases often include errors like misspellings and inaccurate descriptions of video content, as shown in Figure \ref{Fig2}. Moreover, different annotators may characterize the same video segment in varying ways. A minority might opt for a formal language style, while others might gravitate towards colloquial or slang expressions. This difference in descriptions can inadvertently lead the model to prefer certain types of queries, thus introducing human prejudice and potentially diminishing the model's performance.

\begin{figure}[H]
  
\begin{adjustwidth}{-\extralength}{0cm}
\centering 
\includegraphics[width=.92\linewidth]{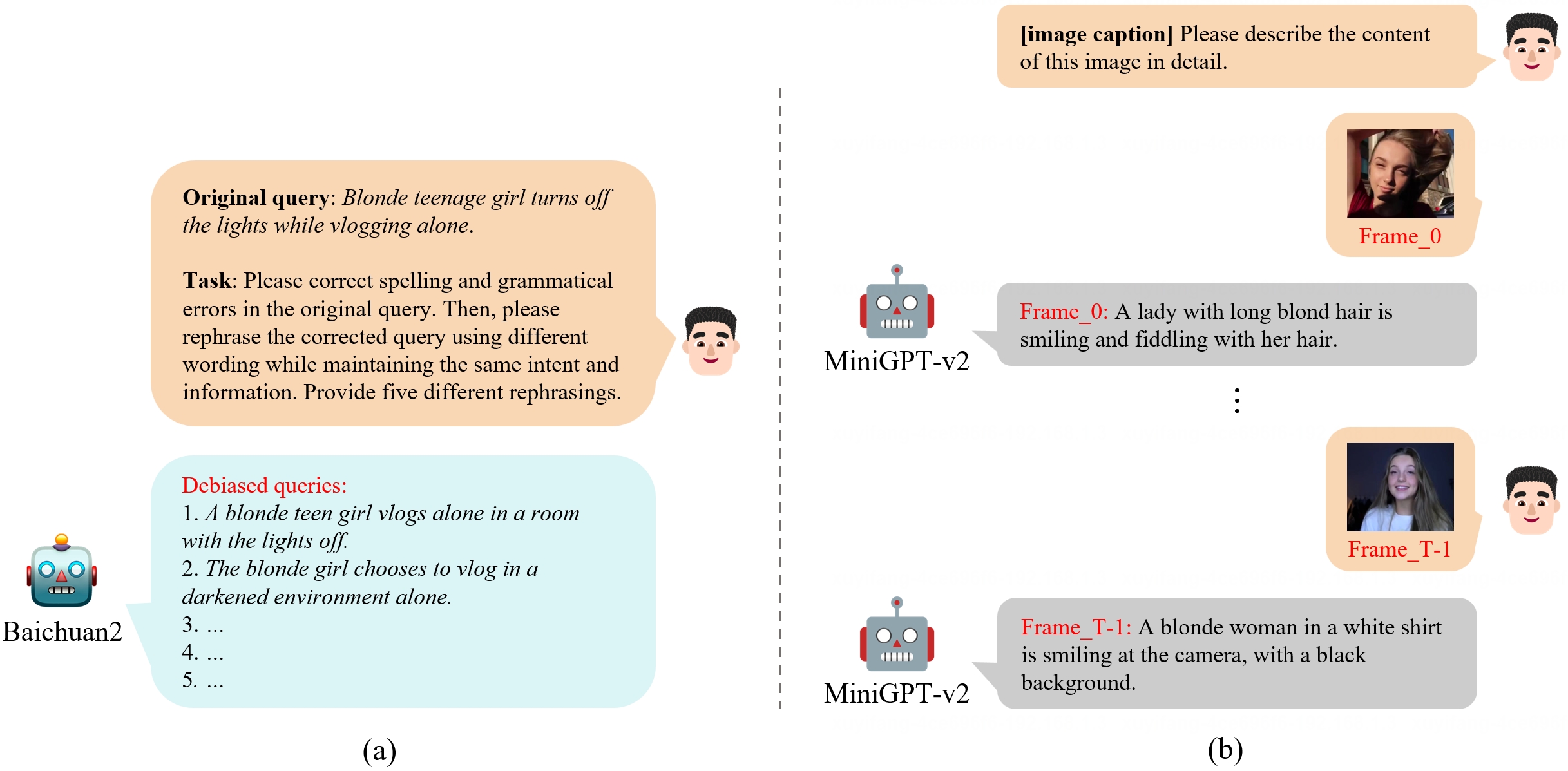}
\end{adjustwidth}
  \caption{(\textbf{a}) An example of query refinement using Baichuan2. (\textbf{b}) An example of image captioning using MiniGPT-v2. The red font employed here is for demonstration purposes only and is not present in the actual code.}
  \label{Fig4}
\end{figure}

\begin{figure}[H]
  
\begin{adjustwidth}{-\extralength}{0cm}
\centering 
\includegraphics[width=0.92\linewidth]{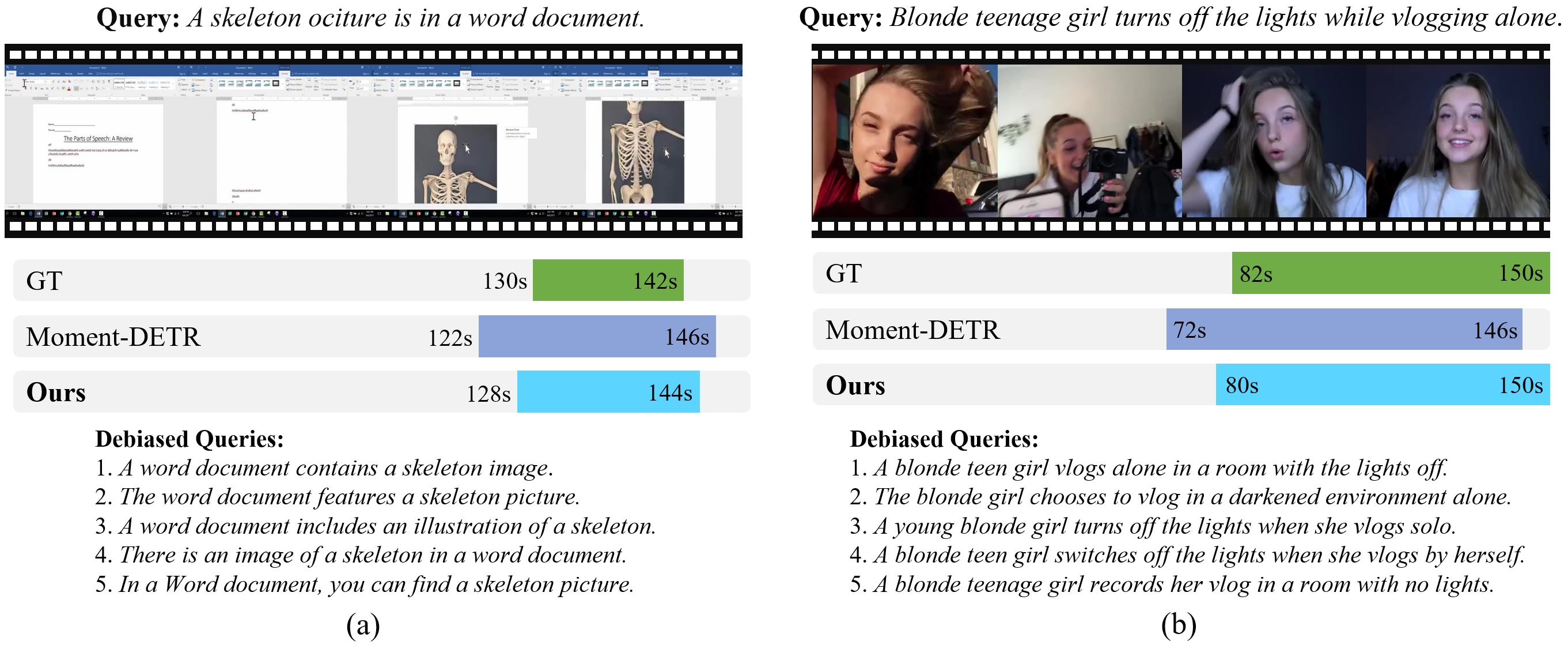}
\end{adjustwidth}
  \caption{{Visualization} 
 of predictions on QVHighlights \textit{{val}} split. ({\textbf{a}) misspelled words. (\textbf{b}) incorrect descriptions.} Our VTG-GPT achieves more precise localization compared to Moment-{DETR} 
 \cite{MomentDETR-2021}, as it can correct errors in the original queries through rewriting and generate debiased queries, thereby facilitating more accurate grounding.}
  \label{Fig5}
\end{figure}

To address the aforementioned challenges, we utilize Baichuan2 to eliminate human biases inherent in original queries, as demonstrated in Figure \ref{Fig4}a. In line with human linguistic comprehension \cite{zhengHumanBehaviorInspired2019}, our first step is to rectify spelling and grammatical inaccuracies in original query $T$, thus producing the corrected version $T_{c}$. We direct GPT with the instruction: \textit{{Please correct spelling and grammatical errors in the original query.}} Subsequently, we instruct Baichuan2 to rewrite $T_{c }$ to remove incorrect descriptions. The corresponding command is \textit{{Please rephrase the corrected query using different wording while maintaining the same intent and information.}} Finally, we generate five semantically similar yet syntactically diverse queries $Q$ to prevent the model from relying on a specific query type. The command for this is \textit{{Provide five different rephrasings.}} Although it is generally advisable to issue only one command per message in GPT dialogues to avoid model errors, as noted in \cite{LLaMA-2023}, we discover in our tests that aggregating all instructions into a single message to GPT proved more effective, as shown in Figure \ref{Fig4}a. It is important to note that the red font is not present in the code. A case involving misspelled words is shown in Figure \ref{Fig5}a, where the incorrectly spelled word ``\textit{{ociture}}'' is corrected to ``\textit{{image}}'' or ``\textit{{picture}}''. Figure \ref{Fig5}b demonstrates a scenario involving a non-existent action, where ``\textit{{turn off the lights}}'' is optimized to ``\textit{{lights off}}'' or ``\textit{{a darkened environment}}'', where ``\textit{{a darkened environment}}'' is more congruent with the original video segment. In short, this debiasing strategy, featuring variations that differ in structure and word choice, deeply explores semantic information and enables the model to process various real-world queries effectively.

\vspace{-9pt}

\subsection{Image Captioning}
\label{sec:image_captioning}
To retrieve corresponding video segments $Seg$ based on the query, traditional zero-shot methods \cite{Zero-shot-VMR-2023, Luo2023ZeroShotVM} initially employ pre-trained multi-modal models \cite{CLIP-2021, InternVideo-2022} for feature extraction from visual and textual modalities. These features are then used to calculate similarities to derive $Seg$. However, our preliminary experiments utilizing CLIP and InternVideo to assess cross-modal similarity, as shown in the upper part of Table \ref{tab1:toy_experiment}, yielded mediocre results. We attribute this to the over-reliance of traditional methods on directly modeling raw frames, which is often influenced by background details, thereby reducing the accuracy of primary content recognition. Some recent works \cite{videomae-2022, SeViLA-2023} suggest that videos contain abundant non-essential information and that translating visual signals into more abstract descriptions can enhance VTG performance.

Inspired by the above research, we incorporate a large multi-modal model (LMM), MiniGPT-v2 \cite{MiniGPT-v2}, to obtain more detailed image descriptions. As demonstrated in \mbox{Figure \ref{Fig4}b}, our initial instruction to MiniGPT-v2 is \textit{{[image caption] Please describe the content of this image in detail.}}, where \textit{{[]}} emphasizes the task to be performed. Subsequently, we sequentially send frames in video $V$ to MiniGPT-v2, which provides us with detailed captions $C \in \mathbb{R}^{N_v \times L_c}$. Following this, we use the CLIP text encoder (CLIP-T) to extract linguistic features from $C$ and $Q$ and calculate their similarities. As illustrated in the third row of Table \ref{tab1:toy_experiment}, the results are surprisingly effective, achieving significant gains with this straightforward approach. We ascribe this to the LMM's focus on capturing key image content, thereby reducing irrelevant background interference and enhancing semantic similarities between queries and frames. For instance, the last frame in Figure \ref{Fig4}b, depicting ``\textit{{A blonde woman in a white shirt is smiling at the camera, with a black background.}}'', is succinctly translated into text, closely matching the query: ``\textit{{The blonde girl chooses to vlog in a darkened environment alone.}}'' semantically.

\begin{table}[H]
\caption{{Preliminary} 
 experiment with different similarity models on QVHighlights \textit{{val}} split, using proposal generator and proposal scorer but without NMS. CLIP-T is short for using {CLIP} 
 \cite{CLIP-2021} text encoder only. Please refer to Section \ref{sec:experimental_settings} for a detailed explanation of the evaluation metrics.}
\label{tab1:toy_experiment}
\newcolumntype{C}{>{\centering\arraybackslash}X}
\begin{tabularx}{\textwidth}{cCCCCC}
\toprule
\multirow{2.4}{*}{{\bf Similarity Models}} & \multicolumn{2}{c}{{\bf R1}} & \multicolumn{3}{c}{{\bf mAP}} \\ \cmidrule(lr){2-3} \cmidrule(ll){4-6}
 & {\bf @0.5} & {\bf {@0.7} 
} & {\bf {@0.5}} & {\bf {@0.75}} & {\bf Avg.} \\ 
\midrule
CLIP \cite{CLIP-2021}  & 45.59 & 26.03 & 45.56 & 23.14 & 24.91 \\
InternVideo \cite{InternVideo-2022} & 49.13 & 32.49 & 48.65 & 25.82 & 26.94 \\
\midrule
CLIP-T \cite{CLIP-2021} & 52.85 & 34.82 & 48.07 & 28.05 & 28.29 \\
RoBERTa \cite{RoBERTa-2019} & \textbf{{54.99}} & 37.58 & 53.77 & 29.18 & 30.15 \\
\textbf{{Sentence-BERT} 
} \cite{Sentence-BERT-2019} & 54.26 & \textbf{{38.45}} & \textbf{{53.96}} & \textbf{{29.25}} & \textbf{{30.38}} \\ 
\bottomrule
\end{tabularx}
\end{table}

\subsection{Proposal Generation}
\label{sec:proposal_generation}

\textbf{{Computing query-frame similarity.} 
} 
In Section \ref{sec:image_captioning}, we have articulated the significance of image captioning within VTG-GPT and employed CLIP-T to model debiased queries $Q$ and image captions $C$ within the textual domain. Subsequently, taking into account CLIP-T, as a multi-modal model, does not outperform specialized language models in NLP (natural language processing) tasks, as outlined in previous research \cite{Survey-LLMs-2023}. Therefore, we explore the use of a language-specific model. We opt for RoBERTa \cite{RoBERTa-2019} (Sentence-BERT~\cite{Sentence-BERT-2019}) to extract normalized pooling features of $Q \in \mathbb{R}^{N_q \times L_t}$ and $C \in \mathbb{R}^{N_v \times L_c}$, denoted as $f_q \in \mathbb{R}^{N_q \times d}$ and $f_c \in \mathbb{R}^{N_v \times d}$, respectively, where $d$ represents the dimensionality. We then compute the cosine scores between $f_q$ and $f_c$ as similarities  $S_s \in \mathbb{R}^{N_q \times N_v}$:
\begin{equation}
S_{s} = \cos(f_q, f_c) = \frac{f_q \cdot f_c}{\|f_q\| \|f_c\|}
\end{equation}
{As} 
 demonstrated in rows four to five of Table \ref{tab1:toy_experiment}, the leverage of expert NLP models yielded significant improvements, which also validates the viewpoints presented in the report \cite{Survey-LLMs-2023}.

\textbf{{Proposal generator.}} 
After obtaining query-frame similarity scores $S_s$, we move towards generating temporal proposals $P \in \mathbb{R}^{N_p \times 2}$. A straightforward method would be to apply a fixed threshold, considering frames with similarity scores exceeding this threshold as potential start or end timestamps. However, each query-video pair exhibits a unique similarity distribution. To adaptively obtain proposals, we introduce a dynamic mechanism within our devised proposal generator. For clarity, we denote the similarity between the \textit{i}-th debiased query $Q^i$ and video $V$ as $S_{s}^{i} \in \mathbb{R}^{N_v}$, and the similarity between $Q^i$ and the \textit{j}-th frame in $V$ as $S_{s}^{i,j} \in \mathbb{R}^{1}$. 

To be specific, the generator begins by computing a histogram of $S_{s}^{i}$ with $N_b$ bins. It then selects the bins containing the top $k$ highest similarities as the dynamic threshold $\theta$:
\begin{equation}
\theta = \text{top\_k}(S_{s}^{i}, N_b, k),
\end{equation}
where $N_b$ and $k$ are hyperparameters. For their specific values, please refer to the implementation details (Section \ref{sec:experimental_settings}) and ablations (Section \ref{sec:ablations}). Next, we iteratively assess each frame; if $S_{s}^{i,j}$ exceeds $\theta$, its corresponding timestamp is considered the proposal's starting point. When more than $\lambda$ consecutive frames are all lower than $\theta$, the last frame with a similarity greater than $\theta$ is marked as the end timestamp of this proposal. Here, $\lambda$ denotes the continuity threshold. Finally, we produce proposals for all debiased queries in the same video using this process to form final temporal proposals $P \in \mathbb{R}^{N_p \times 2}$ (representing potentially relevant video segments).

\subsection{Post-Processing}
\label{sec:post_processing}
\textbf{{Proposal scorer.}} In Section \ref{sec:proposal_generation}, we generate a set of temporal proposals $P$ through our designed proposal generators. To identify the most fitting video segments from $P$, it is essential to compute and rank each proposal's confidence score. Intuitively, a straightforward approach could be averaging the similarity scores for each frame within a proposal, or only considering frames exceeding dynamic threshold $\theta$. However, these methods overlook the impact of proposal length on their scoring. In our experiments, we observe that within certain ground-truth segments containing scene transitions, the similarity of some frames significantly exceeded that of adjacent frames. This led to an excessively high dynamic threshold, resulting in the predicted segments being truncated or fragmented. To address this issue, we develop a length-aware scoring mechanism for proposals, encouraging the model to generate longer segments. Specifically, the evaluation of each proposal considers both its duration and the query-frame similarity, and the final score of each proposal $S_f \in \mathbb{R}^{N_p}$ is calculated as follows:
\begin{equation}
S_f = \alpha \times S_l + (1 - \alpha) \times S_{s},
\end{equation}
where $S_l = L_p / L_n$. Here, $L_p$ represents the count of frames within a proposal exceeding $\theta$, and $L_n$ denotes the total number of frames exceeding $\theta$ across the entire video. The balancing coefficient $\alpha$ is adjustable to optimize for the influence of length and similarity in the final score calculation.

\textbf{{NMS.}} 
In the final stage, considering that multiple debiased queries will produce numerous overlapping proposals, we employ non-maximum suppression (NMS) to reduce redundant overlaps and derive the final predicted video segments $Seg \in \mathbb{R}^{N_s \times 2}$:
\begin{equation}
Seg = \text{NMS}(P, S_f, \mu),
\end{equation}
where segments exceeding the intersection over union (IoU) threshold $\mu$ are selectively eliminated. This method ensures that only the most representative and distinct video segments are retained, enhancing the accuracy and relevance of our VTG-GPT output.







\section{Experiments}
\label{sec:experiments}

\subsection{Experimental Settings}
\label{sec:experimental_settings}
\textbf{{Datasets.}} To demonstrate the superiority and effectiveness of our proposed tuning-free VTG-GPT framework, we conduct extensive experiments on three publicly available datasets: QVHighlights \cite{MomentDETR-2021}, Charades-STA \cite{Charades-STA-dataset-2017}, and ActivityNet-Captions \cite{ActivityNet-Captions-dataset-2017}, as these datasets encompass diverse types of videos. 
\textbf{{QVHighlights}} consists of 10,148 distinct YouTube videos, each accompanied by human annotations that include a textual query, a temporal segment, and frame-level saliency scores. Here, the saliency scores serve as the output for the highlight detection (HD) task, quantifying the relevance between a query and its corresponding frames. QVHighlights encompasses a wide array of themes, ranging from daily activities and travel in everyday vlogs to social and political events in news videos. For evaluation, Moment-DETR \cite{MomentDETR-2021} allocates 15\% of the data for validation and another 15\% for testing, with consistent data distribution across both sets. Due to limitations on the online test {server} 
 (\url{https://codalab.lisn.upsaclay.fr/competitions/6937}, {accessed on 1 September 2023}
) allowing a maximum of five submissions, all our ablation studies are conducted on the validation split. 
\textbf{{Charades-STA}}, derived from the original Charades \cite{Charades-dataset-2016} dataset, includes 9848 videos of human indoor activities, accompanied by 16,128 annotations. For this dataset, a standard split of 3720 annotations is specifically designated for testing.
\textbf{{ActivityNet-Captions}}, built upon the raw ActivityNet \cite{ActivityNet-dataset-2015} dataset, comprises 19,994 long YouTube videos from various domains. Since the test split is reserved for competitive evaluation, we follow the setup used in 2D-TAN \cite{2D-TAN-2020}, utilizing 17,031 annotations for testing.

\textbf{{Metrics.}} 
To effectively evaluate performance on VTG, we employ several metrics, including Recall-1 at Intersection over Union (IoU) thresholds (R1@$m$), mean average precision (mAP), and mean IoU (mIoU). R1@$m$ measures the percentage of queries in the dataset where the highest-scoring predicted segment has an IoU greater than $m$ with the ground truth. mIoU calculates the average IoU across all test samples. For a fair comparison, our results on the QVHighlights dataset report R1@$m$ with $m$ values of 0.5 and 0.7, mAP at IoU thresholds of 0.5 and 0.75, and the average mAP across multiple IoU thresholds [0.5:0.05:0.95]. For the Charades-STA dataset, we report R1@$m$ for $m$ values of 0.3, 0.5, and 0.7, along with mIoU. Finally, we employ mAP and HIT@1 to evaluate the results of HD, thereby measuring the query-frame relevance. Here, HIT@1 represents the accuracy of the highest-scoring frame.

\textbf{{Implementation details.}}
To mitigate video information redundancy, we downsample QVHighlights and Charades-STA datasets to a frame rate of 0.5 per second. Considering the extended duration of videos in the ActivityNet-Captions, we extract one frame every three seconds. In the image captioning stage, we utilize MiniGPT-v2 \cite{MiniGPT-v2} based on the LLaMa-2-Chat-7B \cite{LLaMa-2-2023}. For query debiasing, we employ Baichuan2-7B-Chat \cite{Baichuan-2-2023}, also based on LLaMa-2~\cite{Baichuan-2-2023}, generating five debiased queries ($N_q = 5$) per instance. The temperature coefficients for MiniGPT-v2 and Baichuan2 are set at 0.1 and 0.2, respectively. Drawing from the preliminary experiments in Section \ref{sec:proposal_generation}, we select Sentence-BERT \cite{Sentence-BERT-2019} as our similarity model to evaluate query-frame correlations using cosine similarity. The histogram in our proposal generator is configured with ten bins ($N_b$), with a selection of the top eight values ($k = 8$) and a continuity threshold $\lambda = 6$. During the post-processing phase, the balance coefficient ($\alpha$) in the proposal scorer is set to 0.5, and the IoU threshold ($\mu$) for non-maximum suppression (NMS) is determined at 0.75. All pre-processing and experiments are conducted on eight NVIDIA RTX 3090 GPUs. It is important to note that our VTG-GPT is purely inferential, involving no training phase. 

\subsection{Comparisons to the State-of-the-Art}
In this section, we present a comprehensive comparison of our VTG-GPT with state-of-the-art (SOTA) methods in VTG. Firstly, we disclose results on the QVHighlights validation and test splits, as shown in Table \ref{Table1}. The approaches are categorized into fully supervised (FS), weakly supervised (WS), unsupervised (US), and zero-shot (ZS) methods. Notably, VTG-GPT significantly outperforms the previous SOTA zero-shot model (\mbox{Diwan et al. \cite{Zero-shot-VMR-2023}}), demonstrating substantial improvements across five metrics. Specifically, R1@0.7 saw an increase of +7.49 and mAP@0.5 improved by +7.23. Remarkably, VTG-GPT also vastly exceeds all WS methods. Most impressively, our approach surpasses the FS baseline (Moment-DETR \cite{MomentDETR-2021}) in most metrics, even achieving competitive performance compared with FS methods. Unlike these methods, VTG-GPT requires only a single inference pass, eliminating the need for training data and resources.

\begin{table}[H] 
\centering
\caption{{Performance} 
 comparison on QVHighlights \textit{{test}} and \textit{{val}} split. FS means fully-supervised method, WS means weakly supervised, and ZS means zero-shot.}
\label{Table1}
\begin{adjustbox}{max width=\linewidth}
\begin{tabular}{@{}ccccccccccccc@{}} 
\toprule
\multirow{3.5}{*}{{\bf Method}} & \multirow{3.5}{*}{{\bf Year}} & \multirow{3.5}{*}{{\bf Setup}} & \multicolumn{5}{c}{{\bf QVHighlights test}} & \multicolumn{5}{c}{{\bf QVHighlights val}} \\
 \cmidrule(lr){4-8} \cmidrule(ll){9-13} 
 &  &  & \multicolumn{2}{c}{{\bf R1}} & \multicolumn{3}{c}{{\bf mAP}} & \multicolumn{2}{c}{{\bf R1}} & \multicolumn{3}{c}{{\bf mAP}} \\ 
 \cmidrule(lr){4-5} \cmidrule(lr){6-8} \cmidrule(lr){9-10} \cmidrule(ll){11-13}
 &  &  & {\bf @0.5} & {\bf @0.7} & {\bf @0.5} & {\bf @0.75} & {\bf Avg.}& {\bf @0.5} & {\bf @0.7} & {\bf @0.5} & {\bf @0.75} & {\bf Avg.} \\ 
\midrule
Moment-DETR \cite{MomentDETR-2021} & 2021 & FS & 52.89 & 33.02 & 54.82 & 29.40 & 30.73 & 53.94 & 34.84 & - & - & 32.20 \\
LLaViLo \cite{LLaViLo-2023} & 2023 & FS & 59.23 & 41.42 & 59.72 & - & 36.94 & - & - & - & - & - \\
UMT \cite{UMT-2022} & 2022 & FS & 56.23 & 41.18 & 53.83 & 37.01 & 36.12 & - & - & - & - & 37.79  \\
MH-DETR \cite{MH-DETR-2023} & 2023 & FS & 60.05 & 42.48 & 60.75 & 38.13 & 38.38 & 60.84 & 44.90 & 60.76 & 39.64 & 39.26 \\
QD-Net \cite{QD-Net-2023} & 2023 & FS & 62.32 & 45.61 & 63.15 & 42.05 & 41.46 & 61.71 & 44.76 & 61.88 & 39.84 & 40.34 \\
EaTR \cite{jangKnowingWhereFocus2023} & 2023 & FS & - & - & - & - & - & 61.36 & 45.79 & 61.86 & 41.91 & 41.74 \\
\midrule
CNM \cite{CNM-2022} & 2022 & WS & 14.11 & 3.97 & 11.78 & 2.12 & - & - & - & - & - & - \\
CPL \cite{CPL_2022_CVPR} & 2022 & WS & 30.72 & 10.75 & 22.77 & 7.48 & - & - & - & - & - & - \\
CPI \cite{CPI-2023} & 2023 & WS & 32.26 & 11.81 & 23.74 & 8.25 & - & - & - & - & - & - \\
\midrule
UniVTG \cite{UniVTG-2023} & 2023 & ZS & 25.16 & 8.95 & 27.42 & 7.64 & 10.87 & - & - & - & - & - \\
Diwan et al. \cite{Zero-shot-VMR-2023} & 2023 & ZS & - & - & - & - & -& 48.33 & 30.96 & 46.94 & 25.75 & 27.96  \\
\textbf{{VTG-GPT} 
} (Ours) & 2023 & ZS & \textbf{{53.81}} & \textbf{{38.13}} & \textbf{{54.13}} & \textbf{{29.24}} & \textbf{{30.50}} & \textbf{{54.26}} & \textbf{{38.45}} & \textbf{{54.17}} & \textbf{{29.73}} & \textbf{{30.91}}  \\
\bottomrule
\end{tabular}
\end{adjustbox}
\end{table}

Subsequently, we report the performance on the Charades-STA test set and ActivityNet-Captions test set in Table \ref{Table2}. In Charades-STA, VTG-GPT surpasses the SOTA zero-shot method (Luo et al. \cite{Luo2023ZeroShotVM}) with a +5.81 increase in R1@0.7 and a +1.89 improvement in mIoU. Furthermore, VTG-GPT significantly outperforms the best US method (\mbox{Kim et al. \cite{LFT-VG-2023}}) across all metrics. However, on the ActivityNet-Captions dataset, our method falls slightly behind Luo et al. in two metrics, which we attribute to the high downsampling rate used for this dataset. Moreover, VTG-GPT approaches the performance of the fully supervised Moment-DETR, validating its capacity to handle diverse and complex video contexts without any training or fine-tuning. This underscores the robustness and adaptability of VTG-GPT in zero-shot VTG scenarios, demonstrating its potential as a versatile and efficient tool for video understanding.

\begin{table}[H] 
\centering
\caption{{Performance} 
 comparison on Charades-STA \textit{test} split and ActivityNet-Captions \textit{test} split. Where FS means fully-supervised setting, WS means weakly-supervised, US means unsupervised, and ZS means zero-shot.}
\label{Table2}
\begin{adjustbox}{max width=\linewidth}
\begin{tabular}{@{}cccccccccccc@{}}
\toprule
\multirow{2.5}{*}{{\bf Method}} & \multirow{2.5}{*}{{\bf Year}} & \multirow{2.5}{*}{{\bf Setup}} & \multicolumn{4}{c}{{\bf Charades-STA}} & \multicolumn{4}{c}{{\bf ActivityNet-Captions}} \\ 
\cmidrule(lr){4-7}  \cmidrule(ll){8-11}     
 &  &  &  {\bf R1@0.3} & {\bf R1@0.5} & {\bf R1@0.7} & {\bf mIoU} & {\bf R1@0.3} & {\bf R1@0.5} & {\bf R1@0.7} & {\bf mIoU} \\ 
\midrule
2D-TAN \cite{2D-TAN-2020} & 2020 & FS & 57.31 & 45.75 & 27.88 & 41.05 & 60.32 & 43.41 & 25.04 & 42.45  \\
Moment-DETR \cite{MomentDETR-2021} & 2021 & FS & 65.83 & 52.07 & 30.59 & 45.54 & - & - & - & - \\
VDI \cite{VDI-2023} & 2023 & FS & - & 52.32 & 31.37 & - & - & 48.09 & 28.76 & - \\
\midrule
CNM \cite{CNM-2022} & 2022 & WS  & 60.04 & 35.15 & 14.95 & 38.11 & 55.68 & 33.33 & 13.29 & 37.55\\
CPL \cite{CPL_2022_CVPR} & 2022 & WS  & 65.99 & 49.05 & 22.61 & 43.23 & 55.73 & 31.37 & 13.68 & 36.65\\
Huang et al. \cite{Huang_2023_CVPR} & 2023 & WS & 69.16 & 52.18 & 23.94 & 45.20 & 58.07 & 36.91 & - &  41.02 \\
\midrule
PSVL \cite{PSVL-2021} & 2021 & US & 46.47 & 31.29 & 14.17 & 31.24 & 44.74 & 30.08 & 14.74 & 29.62 \\
Gao et al. \cite{gaoLearningVideoMoment2022} & 2021 & US & 46.69 & 20.14 & 8.27 & - & 46.15 & 26.38 & 11.64 & - \\
DSCNet \cite{DSCNet-2022} & 2022 & US & 44.15 & 28.73 & 14.67 & - & 47.29 & 28.16 & - & - \\
PZVMR \cite{PZVMR-2022} & 2022 & US & 46.83 & 33.21 & 18.51 & 32.62 & 45.73 & 31.26 & 17.84 & 30.35 \\
Kim et al. \cite{LFT-VG-2023} & 2023 & US & 52.95 & 37.24 & 19.33 & 36.05 & 47.61 & 32.59 & 15.42 & 31.85 \\
\midrule
UniVTG \cite{UniVTG-2023} & 2023 & ZS  & 44.09 & 25.22 & 10.03 & 27.12 & - & - & - & - \\
Luo et al. \cite{Luo2023ZeroShotVM} & 2023 & ZS  & 56.77 & 42.93 & 20.13 & 37.92 & \textbf{{48.28}} & 27.90 & 11.57 & \textbf{{32.37}}\\
\textbf{{VTG-GPT} 
} (Ours) & 2023 & ZS & \textbf{{59.48}} & \textbf{{43.68}} & \textbf{{25.94}} & \textbf{{39.81}} & 47.13 & \textbf{{28.25}} & \textbf{{12.84}} & 30.49 \\
\bottomrule
\end{tabular}
\end{adjustbox}
\end{table}

To qualitatively validate the effectiveness of our VTG-GPT model, we present visual comparisons of grounding results from the Ground-Truth (GT), Moment-DETR, and VTG-GPT in Figure \ref{Fig5}. Observations indicate that the tuning-free VTG-GPT achieves more precise localization than the supervised Moment-DETR. The primary reason is that Moment-DETR relies solely on the original queries, which contain human-annotated errors, thus failing to fully align with the video's semantic information. In contrast, VTG-GPT can correct erroneous queries and reduce the bias introduced by human annotations, leading to more accurate grounding. To be more specific, in Figure \ref{Fig5}a, our model detects a spelling mistake in the query, where ``\textit{{ociture}}'' is corrected to ``\textit{{image}}'' or ``\textit{{picture}}''. In Figure \ref{Fig5}b, the action ``\textit{{turns off}}'' is refined to terms more congruent with the video context, such as ``\textit{{lights off}}'', ``\textit{{darkened environment}}'', and ``\textit{{no lights}}''. Additionally, the five rephrasings of each original query, in contrast to the original phrasing, exhibit more flexible grammatical structures, enabling the text encoder to comprehensively capture the semantic information of the original query.

\subsection{Ablation Studies}
\label{sec:ablations}
To demonstrate the effectiveness of each module within our VTG-GPT framework, we perform in-depth ablation studies on the QVHighlights dataset.

\textbf{{Effect of debiased query.}} 
Firstly, we report saliency scores used to evaluate query-frame correlation. As delineated in Table \ref{tab:HD}, row three corresponds to VTG-GPT without debiasing, where we directly employ the similarity generated by Sentence-BERT as the saliency scores. Conversely, row four is VTG-GPT with debiasing, wherein we average the similarity of five debiased queries as saliency scores. The comparison reveals that row four significantly outperforms row three, demonstrating the efficacy of our debiasing strategy in mitigating human biases inherent in the original queries. Furthermore, comparing row two (UMT~\cite{UMT-2022}) and row four, our VTG-GPT achieves a notable increase in HIT@1, recording a score of 62.29 (+2.3). This enhancement underscores VTG-GPT's superior reasoning capabilities in discerning challenging cases, affirming the value of our debiasing approach in refining model performance.

\begin{table}[H]
\caption{{Comparison} 
 of video highlight detection (HD) on QVHighlights \textit{{val}} split. VG is the abbreviation of very good. {\ding{51} and \ding{55} respectively represent the use and non-use of debiased queries.}
}
\label{tab:HD}
\newcolumntype{C}{>{\centering\arraybackslash}X}
\begin{tabularx}{\textwidth}{cCCCC}
\toprule
\multirow{2.5}{*}{{\bf Methods}} & \multirow{2.5}{*}{{\bf Setup}} & \multirow{2.5}{*}{{\bf Debiasing}} & \multicolumn{2}{c}{{\bf HD (}\boldmath{$\geq$}{\bf VG)}} \\ 
\cmidrule(l){4-5} 
 &  &  & {\bf mAP} & {\bf HIT@1} \\ 
\midrule
Moment-DETR \cite{MomentDETR-2021} & FS & \ding{55} & 35.69 & 55.60 \\
UMT \cite{UMT-2022} & FS & \ding{55} & \textbf{{38.18} 
} & 59.99 \\ 
\midrule
\multirow{2}{*}{\textbf{{VTG-GPT}}} & \multirow{2}{*}{ZS}  & \ding{55} & 34.84 & 60.48 \\
 & & \ding{51} & 36.08 & \textbf{{62.29}} \\ 
\bottomrule
\end{tabularx}
\end{table}

\begin{figure}[H]
 
\begin{adjustwidth}{-\extralength}{0cm}
\centering 
 \includegraphics[width=1\linewidth]{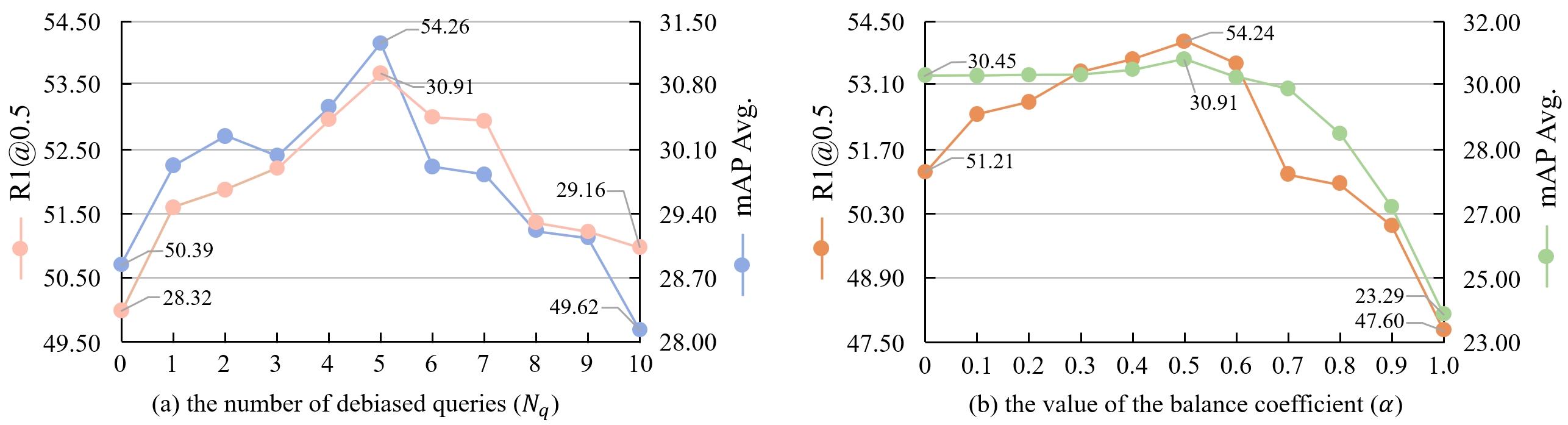}
\end{adjustwidth}
  \caption{Ablation experiments on the QVHighlights \textit{val} split focus on R1@0.5 and mAP Avg. (\textbf{a})~Utilizing debiased queries can enhance model performance, yet increasing the number of debiased queries ($N_q$) does not always lead to better results. The model achieves optimal performance when $N_q$ is set to 5. (\textbf{b}) In the proposal scorer, proposal length significantly impacts the final outcomes, with the model performing optimally when $\alpha=0.5$.}
  \label{Fig6}
\end{figure}

Then, we investigate the impact of different numbers of debiased queries ($N_q$) generated by Baichuan2 on the performance of the VTG-GPT model. Our findings, as depicted in Figure \ref{Fig6}a, indicate that the model achieves optimal results when utilizing five debiased queries ($N_q = 5$). Compared to using solely the original biased query, implementing five debiased queries resulted in a notable increase in R1@0.5 to 54.26 (+3.87) and an improvement in mAP Avg. to 30.91 (+2.59). This evidence suggests that removing bias from queries significantly enhances the model's accuracy. However, an intriguing observation emerged: the performance metrics decline when $N_q$ exceeds 5. This pattern suggests that excessive rephrasing does not continually yield improvements, likely due to the finite number of synonymous rewrites and syntactic variations available to maintain the original intent of the query. Over-rephrasing can introduce irrelevant content, deviating from the semantic intent of the raw query, and potentially diminishing model performance. This finding underscores the critical need to balance the number of query rewrites, ensuring that debiased queries capture a spectrum of semantic nuances while retaining the essence of the original query. Future research should focus on developing advanced query debiasing techniques to enhance this equilibrium.

\textbf{{LLMs and LMMs.}} 
In Table \ref{tab: different LLMs}, we evaluate the capabilities of LLMs (LLaMA-v2 \cite{LLaMa-2-2023} and Baichuan2 \cite{Baichuan-2-2023}), alongside LMMs (MiniGPT-4 \cite{MiniGPT4-2023} and MiniGPT-v2 \cite{MiniGPT-v2}) in handling biased queries and generating image captions. A comparison between rows two and five reveals that Baichuan2 outperforms LLaMa-v2, since it is trained on a more diverse dataset and tasks based on LLaMa-v2, enhancing its sentence rewriting capabilities. As illustrated in row three, MiniGPT-v2, also developed on the foundations of LLaMa-v2, shows moderate results in text dialogue. Comparing rows four and five, we observe an improvement in the performance of MiniGPT-v2 over MiniGPT-4. Overall, the results suggest that the integration of Baichuan2 for query debiasing combined with MiniGPT-v2 for image captioning emerges as the most effective strategy. This effectiveness stems from their complementary capabilities: Baichuan2 excels in handling complex multi-turn text dialogues, while MiniGPT-v2 is adept at detailed multimodal dialogues. This synergy maximizes the text comprehension ability of LLMs and the video understanding capacity of LMMs, thereby enhancing the overall performance of our framework.

\begin{figure}[H]
  \includegraphics[width=0.55\linewidth]{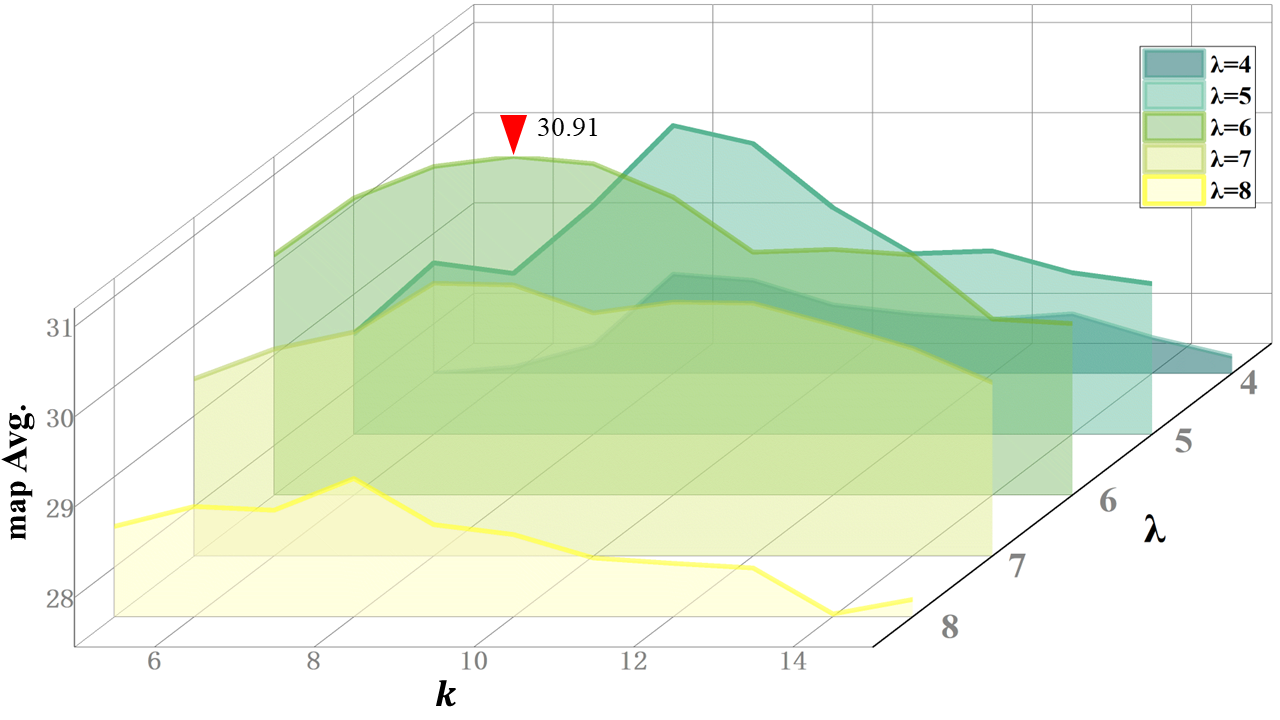}
  \caption{{Ablation} 
 experiments for top-$k$ and continuity threshold ($\lambda$) in proposal generator on QVHighlights \textit{val} split. When $k=8$ and $\lambda=6$, the model achieves the best performance (red triangle).}
  \label{Fig6-topk and λ}
\end{figure}

\textbf{{Proposal generator.}} 
In our study, top-$k$ and the continuity threshold $\lambda$ within the proposal generator play a critical role. The parameter $k$, acting as a count threshold in our dynamic mechanism, directly influences the identified length of relevant proposals. In contrast, $\lambda$ determines the number of irrelevant consecutive frames. To optimize these parameters, we conducted a series of ablation experiments on the proposal generator, as illustrated in Figure \ref{Fig6-topk and λ}. The visualized results indicate that a combination of $k = 8$ and $\lambda = 6$ yields the most favorable outcomes. This specific pairing strikes a balance between segment length and threshold sensitivity. It skillfully avoids the drawbacks of excessively low thresholds, which could incorporate irrelevant frames into prediction results. Simultaneously, it averts the "tolerance trap" where an overly high number of discontinuous frames makes it difficult to determine when the segment ends.

\begin{table}[H]
%
\caption{Ablation study of different LLMs and LMMs (Large Multi-modal Models) on QVHighlights \textit{{val}} split.}
\label{tab: different LLMs}
\newcolumntype{C}{>{\centering\arraybackslash}X}
\begin{tabularx}{\textwidth}{CCCCCCCCC}
\toprule
{\bf Debiasing} & {\bf Captioning} & {\bf R1@0.5} & {\bf R1@0.7} & {\bf mAP Avg.}\\
\midrule
LLaMA-v2 \cite{LLaMa-2-2023} & MiniGPT-4 \cite{MiniGPT4-2023} & 50.78 & 30.56 & 27.20 \\
LLaMA-v2  & MiniGPT-v2 \cite{MiniGPT-v2} & \textbf{{54.65}} & 34.08 & 30.15 \\
MiniGPT-v2  & MiniGPT-v2 & 50.46 & 29.87 & 27.48 \\
Baichuan2 \cite{Baichuan-2-2023} & MiniGPT-4  & 52.78 & 33.84 & 28.54 \\
\textbf{{Baichuan2} 
}  & \textbf{{MiniGPT-v2}} & 54.26 & \textbf{{38.45}} & \textbf{{30.91}}  \\
\bottomrule
\end{tabularx}
\end{table}
%
%

\vspace{-6pt}

\textbf{{Proposal scorer.}} 
To balance the quality and length of segments, we conduct experiments on our proposal scorer, as shown in Figure \ref{Fig6}b. We explore integrating the length score $S_{l}$ into the scoring mechanism. Initially, without including the length score ($\alpha = 0$), mAP Avg. is 30.45. Upon incorporating $S_{l}$, mAP Avg. peak at 30.91. Similarly, R1@0.5 increases from 51.21 to 54.24, indicating that incorporating a length-based scoring mechanism is crucial for generating the final segment scores.

\textbf{{IoU threshold} $\mu$.} 
Finally, we assess the effectiveness of IoU thresholds $\mu$ in the NMS process, focusing on their role in reducing segment overlap. It is important to note that NMS does not alter the values of R1@0.5 and R1@0.75. Therefore, we report only the mAP metrics in Table \ref{tab:tabel 4-NMS-IOU}. As illustrated in Table \ref{tab:tabel 4-NMS-IOU}, setting $\mu$ to 0.75, compared to not employing NMS ($\mu = 1$), results in an increase of +0.53 in mAP Avg. This increment underscores the significance of eliminating excessively overlapping segments, affirming that reducing such overlaps can notably enhance the model's performance. 

\begin{table}[H]
\caption{Comparison of different IoU thresholds ($\mu$) in NMS on QVHighlights \textit{val} split.}
\label{tab:tabel 4-NMS-IOU}
\newcolumntype{C}{>{\centering\arraybackslash}X}
\begin{tabularx}{\textwidth}{CCCCCC}
\toprule
\multirow{1}{*}{\boldmath{$\mu$}} & {\bf mAP@0.5} & {\bf mAP@0.75} & {\bf mAP Avg.}  \\ 
\midrule
0.6 & 53.71 & 28.48 & 30.06 \\
0.7 & 54.12  & 29.60 & 30.54 \\
\textbf{{0.75} 
} & \textbf{{54.17}} & 29.73 & \textbf{{30.91}} \\
0.8 & 54.02 & \textbf{{29.87}} & 30.68 \\
0.9 & 53.81 & 29.63 & 30.41 \\
1.0 & 53.96 & 29.25 & 30.38 \\
\bottomrule
\end{tabularx}
%
\end{table}

\section{Conclusions}
\label{sec:conclusion}
This paper proposes a tuning-free framework named VTG-GPT for zero-shot video temporal grounding. To minimize the bias from mismatched videos and queries, we employ Baichuan2 for refining human-annotated queries. Recognizing the inherent redundancy in video compared to text, we utilize MiniGPT-v2 to transform visual inputs into more exact descriptions. Moreover, we develop the proposal generator and post-processing to produce temporal segments from debiased queries and image descriptions. Comprehensive experiments validate that VTG-GPT significantly surpasses current SOTA methods in zero-shot settings. Remarkably, it achieves a level of performance on par with supervised~approaches.

\section{Discussion}
\textbf{{Limitations.}} In our study, constrained by computational resources, we downsample frames in the long-video dataset ActivityNet-Captions, which adversely affected performance. Future work should focus on developing a more efficient and rapid GPT model to address this challenge. Moreover, due to the limitations imposed by the context length in video-based GPT, our framework relies solely on image-based GPT, thus needing more temporal information modeling. 

\textbf{{In future work}}, 
we will explore applying video-based GPT (such as VideoChatGPT~\cite{VideoChatGPT-2023}) to enhance the capabilities of zero-shot VTG. In addition, crafting a more efficient module for query debiasing and proposal generation is paramount. Finally, leveraging GPT to implement a zero-shot framework on other data-driven tasks (such as video summarization~\cite{UniVTG-2023}, depth estimation \cite{PFANet-2021, DAFFNet-2021} and transformer diagnosis \cite{jiang2024transient}) is very promising.

\textbf{{Ethical considerations.}} 
Our work is based on open-source LLMs and LMMs which require direct inference without training, thereby reducing the carbon footprint. Additionally, we utilize common and safe prompts, and have not observed the generation of harmful or offensive content by the model.
\vspace{6pt}

\authorcontributions{{Conceptualization, Y.X. and Y.S.; methodology, Y.X.; software, Y.X. and Y.S.; validation, Z.X. and B.Z.; formal analysis, Y.X.; investigation, Y.X.; resources, Y.S.; data curation, Y.X.; writing---original draft preparation, Y.X. and Y.S.; writing---review and editing, Z.X. and B.Z.; visualization, Z.X. and B.Z.; supervision, S.D.; project administration, S.D.; funding acquisition, S.D. All authors have read and agreed to the published version of the manuscript.} 
}

\funding{{This research received no external funding.} 
}

\institutionalreview{{Not applicable.} 
}

\informedconsent{{Not applicable.} 
}

\dataavailability{{Publicly available datasets were analyzed in this study. This data can be found here: 
 \url{https://github.com/YoucanBaby/VTG-GPT}.} 
}


\acknowledgments{{Many} 
 thanks to Youyao Jia for his discussion and help in polishing this paper.}

\conflictsofinterest{The authors declare no conflicts of interest.} 

\begin{adjustwidth}{-\extralength}{0cm}

\reftitle{{References}}




 \PublishersNote{}
\end{adjustwidth}
\end{document}